# Transfer learning using deep neural networks for Ear Presentation Attack Detection: New Database for PAD

Jalil Nourmohammadi Khiarak [*]

[*] Institute of Control and Computation Engineering, Warsaw University of Technology, Warsaw, Poland

## ABSTRACT

Ear recognition system has been widely studied whereas there are just few ear presentation attack detection methods for ear recognition systems, consequently, there is no publicly available ear presentation attack detection (PAD) database. In this paper, we propose a PAD method using pre-trained deep neural network and release a new dataset called Warsaw University of Technology Ear Dataset for Presentation Attack Detection (WUT-Ear V1.0). There is no ear database that is captured using mobile devices. Hence, we have captured more than 8500 genuine ear images from 134 subjects and more than 8500 fake ear images using. We made replay-attack and photo print attack with 3 different mobile devices. Our approach achieves 99.83% and 0.08% for the half total error rate (HTER) and attack presentation classification error rate (APCER), respectively, on the replay-attack database. The captured data is analyzed and visualized statistically to find out its importance and making it a benchmark for further research. The experiments have been found out a secure PAD method for ear recognition system, publicly available ear image, and ear PAD dataset. The codes and evaluation results are publicly available at https://github.com/Jalilnkh/KartalOl-EAR-PAD .

## 1. Introduction

At Studies in biometric explains that whatever you make a high accuracy biometric recognition system, having PAD is necessary on your recognition system [1-3]. Moreover, it is still required to improve ear recognition system with different ideas like using multimodal biometric which is proposed in this paper. Ear biometric has been proved as a stable biometric, cost effective because of being easy to capture, and needs no more hardware.

Recently, lots of comprehensive research papers and surveys have been published on-ear recognition systems which show the importance of ear biometrics for study and investigation [4, 5]. Like the face, ear shape is formed of several parts such as helix, incisura, antitragus, antihelix, lope, tragus etc. These features can be useful for recognition, moreover they show the complexity of ear to be comparable to significant biometrics like face. In contrast, ear biometric is vulnerable in facing with presentation attacks. Ear shape can be easily is stolen by attackers using a camera. Attackers make fake ear photo, artificial ear, video of ear and they fool the ear recognition system. To prevent these kinds of attack, some researches have done recently. It should be noted that there is no research on PAD ear on mobile devices.

In this paper we propose a PAD for ear recognition based on DNN, hence, we have done transfer learning on MobileNetV2 [6] with two different parameters. Despite the fact that MobileNetV2 has less parameter, it has been known as a good DNN for mobile applications.

In the rest of paper, we review the recent related work on ear PAD. Then our proposed method for ear PAD is presented. After that, we describe Multimodal mobile biometric dataset (WUT-Ear V1.0) in details and visualize them with t-distributed stochastic neighbor embedding (T-SNE) method. In the same section, we evaluate our proposed method for ear recognition and PAD with baseline methods and at the end; we conclude what we have done in the paper.

## 2. Literature review

Presentation attack detection (PAD) for ear biometric has been studied recently [7]. In 2018, a small anti-spoofing dataset from the existing dataset called AMI and their own dataset has been made for ear PAD [7]. They have used Image Quality Assessment for feature extraction and in following on an SVM classification was applied to the features to distinguish between real and fake input. Three kinds of attacks have been designed for ear PAD namely; video attack, display attack, and photo attack. Half Total Error Rate (HTER) for their own dataset and AMI dataset were 22.4 and 0 respectively. To our best knowledge, it was the only work that was available at that time. However, some other researchers have published ear PAD methods later on that year. Image quality assessment (IQA) measurements are used in PAD widely. In [8], a no-reference and full-reference IQA methods are used for detecting print attacks. They used 21 full-reference and 4 no-references IQA methods and the outputs are classified by a K-Nearest Neighbor classifier. To test the proposed method, they have used AMI and University of Beira EAR (UBEAR) as a real image and they have made 200 fake images using a printer. The HTER in the best condition is respectively 15.5% and 8.5% for the UBEAR and AMI databases. The accuracy of the method is not enough for a real application biometric system. The fake data is less and just one attack has been considered as attack presentations.

Also, In [9], both full-reference and no-reference IQA methods are utilized to detect attacks on ear biometrics with more experiments on the datasets. They proposed a three-level fusion method including; a) feature extraction using both full-reference and no-reference IQA methods, b) score normalization to move them into the same scale and c) decision level. Printed images on paper are used as attack purposes which are made using AMI and UBEAR datasets. The HTER in the best condition is respectively 1% and 10% for the AMI [10] and UBEAR databases [11]. Moreover, they have shown that CNN-based method doesn't achieve as well accuracy as their proposed method in UBEAR database in which HTER for their methods and CNN-based method 14.5% and 10.0% has been reported respectively. The fake data is less and one attack is considered as attack presentations. As a matter of fact, CNN-based methods need more dataset to get better results. Therefore, the reported accuracy cannot be trustable.

In [12] a dataset is collected for ear PAD, the Lenslet Light Field Ear Artefact Database (LLFEADB), including light field and 2D ear artifact images. The dataset has two subsets; Baseline Set (BS) is formed 268 ear images from 67 users, Lytro ILLUM lenslet light field camera for image capturing, and the images are resized to 192*128 pixels and Extended High-Resolution Set (EHRS) is formed 60 images from 14 users from IST-EURECOM LLFEDB bona fide images and the images are resized to 1152*768 pixels. Four types of attacks are considered; Laptop Attack (MacBook Pro 13'' display device), Tablet Attack (iPad Air2, 9,7''), Mobile Attack 1,2(iPhone 6S and Sony Xperia z2). Conventional 2D Methods (C2DM) and Light Field base Methods (LFM) are used as PAD algorithms. Attack Presentation Classification Error Rate (APCER) for LFM was better than other methods and has achieved 0.0% for all exitance attacks in the dataset. Based on 64-bit Intel PC with a 3.40 GHz processor and 16 GB RAM, MATLAB R2015b, running time was 217ms per image.

## 3. Database

Lack of ear PAD large database based on mobile devices gives us a motivation to collect a database. We have collected a large database for ear verification and more importantly for ear PAD, namely Multimodal mobile biometric dataset (WUT-Ear V1.0).

### 3.1. Ear photo database

WUT-Ear V1.0 database is composed of ear touches and images which were collected from students around the world (namely; Azerbaijan, Afghanistan, Algeria, China, Ecuador, India, Iran, Jordan, Latvia, Mexico, Oman, Poland, Portugal, Spain, Turkey, Vietnam, Uzbekistan). The multi-biometric data has been gathered in wild conditions. It was primarily designed for testing algorithms against presentation attacks, specifically on mobile phones [7]. The database has been acquired from 137 different users, with different ages. For each subject, more than 70 images have been taken.

#### 3.1.1. Data collection methodology

The Samsung Galaxy A7 is used to take images and touch information. To take images flash is used. The distance between a subject's ear and mobile phone is about 15 cm. 35 images from 5 different positions are taken in one session from each side therefore totally there are almost 70 images however some subjects have more than 70. It should be noted that more than 10 subjects have earrings. The positions are: up, down, front, forward, and back.

In addition, all images were preprocessed and numbered with a property name. The size of preprocessed images and raw images were not changed. Approximately, raw images and preprocessed images have 4608*3456 and 1992*3120 pixels, respectively. All images in the database are JPEG format. Two instances of both raw and preprocessed subject's images from WUT-Ear V1.0 are shown in Fig. 1 and Fig. 2 respectively. Moreover, the details of the database have been illustrated in Table 1.

#### 3.1.2. Some statistics

WUT-Ear V1.0 is a dataset therefore, since we are not going to use 8638 extracted features from the images in the calculations, we take a subset of the images which the images are selected randomly. Hence, to visualize we just used 10 subjects overall 100 images from genuine images. Most of researchers use PCA for image visualization while PCA is used for linear analysis and it is not sufficient to visualize image dataset. To handle the issue, we have done transfer learning on MobilenetV2 with two neurons in embedded layers [13]. In the other word, we added a layer with two neuron and retrained the network. This helps us to follow nonlinearity roles and the results showed that using PCA cannot help us to analysis image datasets sufficiently. Therefore, as it is shown in Fig. 3 (a), despite the fact that we have just two features from a 128*80*3 image, we can show a good distribution of samples on a plot. It should be noted that still the extracted features by DNN method doesn't give us concrete information, therefore, we used an introduced method by Geoffrey Hinton et al. in [14] called T-SNE to have a good visualization of the datasets (Fig. 3 (b)). T-SNE shows a good clustering of samples, it means that the dataset might has high quality to work on it. Samples in Fig. 3 are extracted by the best proposed DNN called Earnet-1 in this paper.

Table 1: Ear Fake dataset details

| Type of attack | Represerntor | Recognition devices | Number of Fake Ears | Total Number of Images |
|---|---|---|---|---|
| Display attack | Dell UltraSharp 32 Ultra HD 4K Monitor | Samsung Galaxy A7 | 2134 | 5062 |
| | | Samsung Galaxy S9 | 2827 | |
| | | Nokia lumia 1020 | 101 | |
| Display attack | SAMSUNG C27JG50QQUX monitor | Samsung Galaxy A7 | 16 | 3411 |
| | | Samsung Galaxy S9 | 2026 | |
| | | Nokia lumia 1020 | 1369 | |
| Photo attack | Brother MFC-9340CDW - multifunction printer | Samsung Galaxy A7 | 189 | 189 |

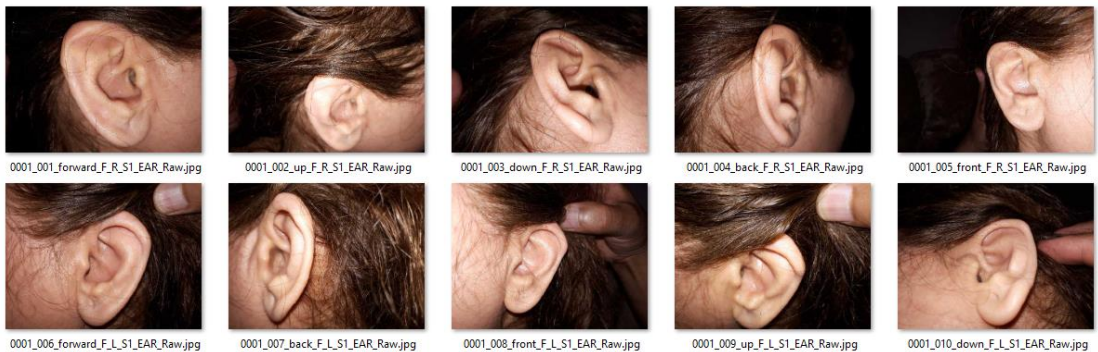

Fig. 1. An example of raw ear images from WUT-Ear V1.0 databases

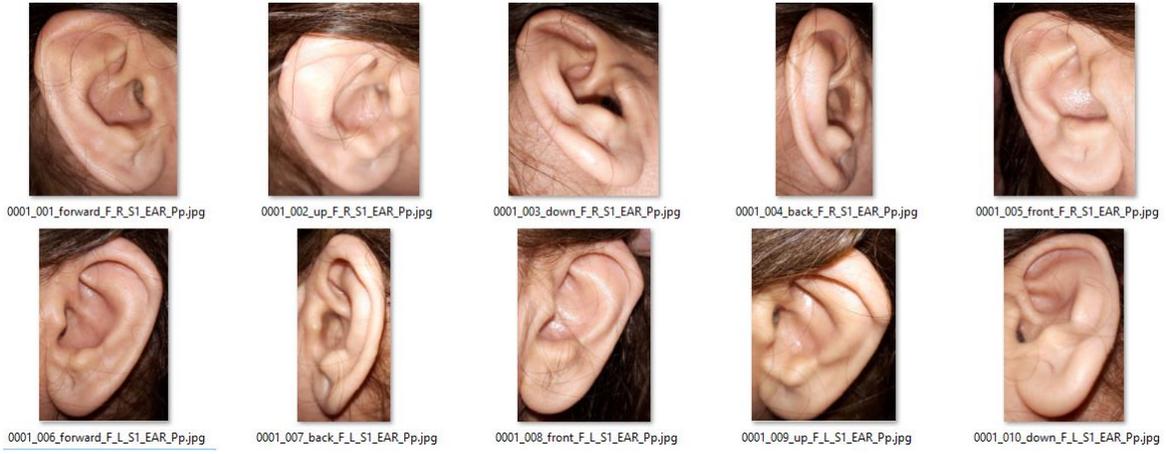

Fig. 2. An example of preprocessed ear images from WUT-Ear V1.0 databases

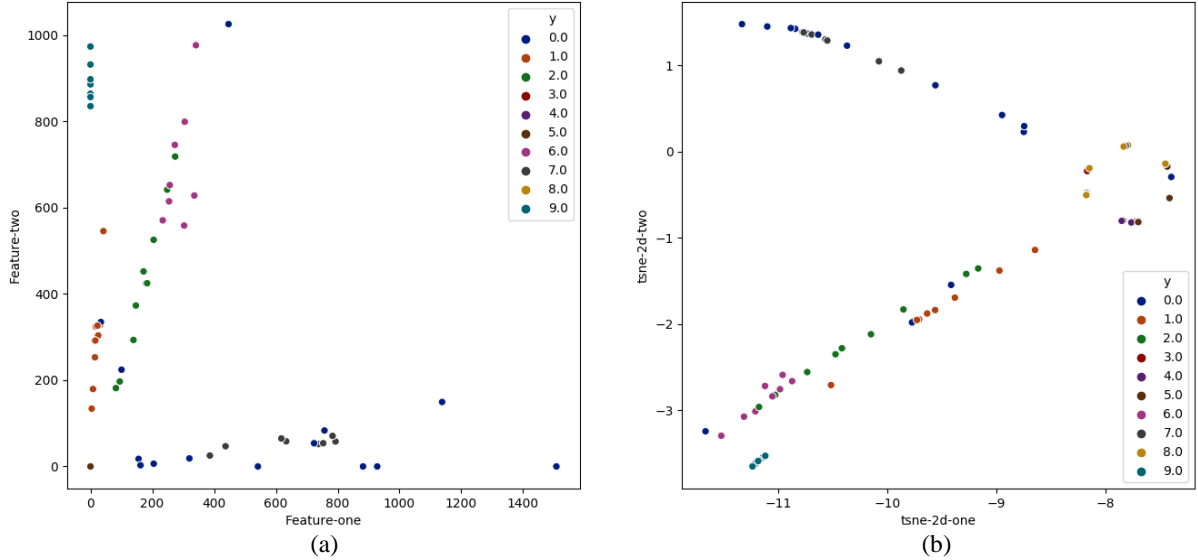

(a)           (b)

Fig. 3: Data visualization using T-SNE, a) a distribution of ear photo in 2D grid, all features are extracted using a PCA. b) a distribution of ear photo in 2D grid, all features are extracted using a T-SNE.

### 3.1.3. Ear pad database

This database is formed of two parts real and fake images which contain more than 8000 real images of 137 subjects and 8662 fake images. Real images have been collected using Galaxy A7 smartphone camera. Fake images have been made with different scenarios. In Fig. 5, we show instance images of spoof and real ears of one subject in our dataset. In the database, there are three different attacks have been made using different sensors for display attack the details of the database have been shown in Table 2.

Table 2: details of Ear Fake Database using different types of representor and recognition device

| Type of attack | Representor | Recognition devices | Number of Fake Ears | Total Number of Images |
|---|---|---|---|---|
| Display attack | Dell UltraSharp 32 Ultra HD 4K Monitor | Samsung Galaxy A7 | 2134 | 5062 |
| | | Samsung Galaxy S9 | 2827 | |
| | | Nokia lumia 1020 | 101 | |
| Display attack | SAMSUNG C27JG50QQUX monitor | Samsung Galaxy A7 | 16 | 3411 |
| | | Samsung Galaxy S9 | 2026 | |
| | | Nokia lumia 1020 | 1369 | |
| Photo attack | Brother MFC-9340CDW - multifunction printer | Samsung Galaxy A7 | 189 | 189 |

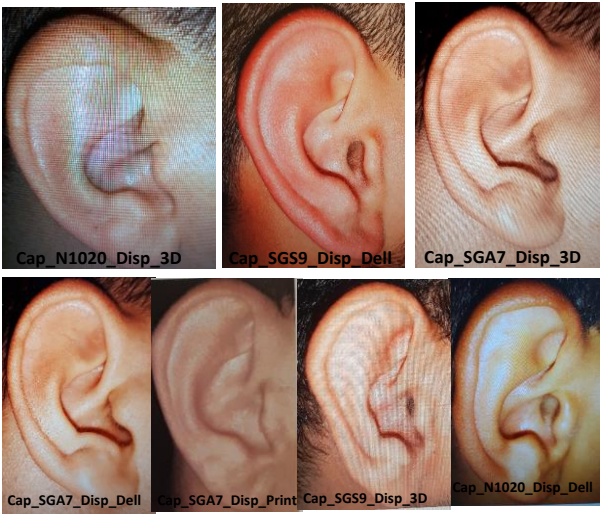

Fig. 4: Ear fake images samples, the written code bottom of each images shows the recognition system and display screen to take photos. For instance, Cap_N1020_Disp_3D means, we used Nokia lumia 1020 as a recognition system to capture photo and SAMSUNG C27JG50QQUX monitor as a display or representor.

### 3.1.3.1. Presentations attack instruments

**Display attack:** Display attack is considered one of the most important attacks in presentation attack detection. In these scenarios, attackers try to unlock the system by showing the fake image instead of having a real enrolled person in the recognition system via a monitor or a smartphone. To make fake images, we used a **Dell UltraSharp 32 Ultra HD 4K Monitor** and **SAMSUNG C27JG50QQUX monitor** which had good quality and are suitable for making fake images. Then a Samsung Galaxy A7, Samsung Galaxy S9, and Nokia lumia 1020 smartphone were used for taking photos of the displayed images.

**Photo attack:** First, we print the genuine photos on A4 glossy paper. For doing this, we used a **Brother MFC-9340CDW - multifunction printer - color Specs** printer which had good quality and is suitable for making fake images. Then a Samsung Galaxy A7 smartphone was used for taking photos of the printed images. The average distances for printed photos are considered as genuine images (~15cm). The sizes of printed photos and the genuine ear photo are the same. The dataset has two suitable advantages: a) all of the images have been captured by the mobile device and it is good for the mobile phone to unlock applications, and b) the printed photos are generated with a good quality printer and the size of the fake ear is the same size as the genuine ear.

### 4. Ear pad methodology

The best solution for having deep learning on the small dataset is to using transfer learning on a pre-trained existing deep neural network (DNN) model. There are two ways to do transfer learning; using the DNN model as a feature extractor without training new task's data and doing fine-tune on the network which can be done on the whole network's layers or some of those layers. Transfer learning may reduce computational resources or training time, training from scratch takes several days or months depend on the amount of data, and avoid overfitting. The proposed ear PAD network (PADNet) follows the MobileNetV2 [6] architecture except for the added top layers. The PADNet architecture is depicted in Fig. 6. In this paper, we use MobileNetV2 which was trained on the ImageNet classification database with 1000 classes. The PADNet modifies compared to the MobileNetv2 are just at the last layers. We removed fully connected layer from MobileNetV2. We fine-tune MobileNetV2 with our collected data for ear PAD.

The MobileNetV2 architecture is a 2D CNN and the image size is determined 224*224*3 pixels as default input. An inverted residual model was used in MobileNetV2 where the residual black has thin bottleneck layers for its output and input. The network has 28 convolutional layers along with 19 residual bottleneck layers and kernel size was always 3*3. Some other parameters have been used such as Rectified Linear Unit (ReLU) as an activation function, batch normalization, and dropout. Moreover, the decision function is sigmoid function, which is better for binary classification.

To fine-tune the MobileNetV2, we design two different versions of MobileNetV2. To see our modification on the network we call it PADNet (It is not a new neural network method. We changed the MobileNetV2 just to make short form and relate it to our research). In PADNet-1 we froze 26 layers and we add three dense layers which are shown in Table 3. We split our classes to two classes real images and fake images.

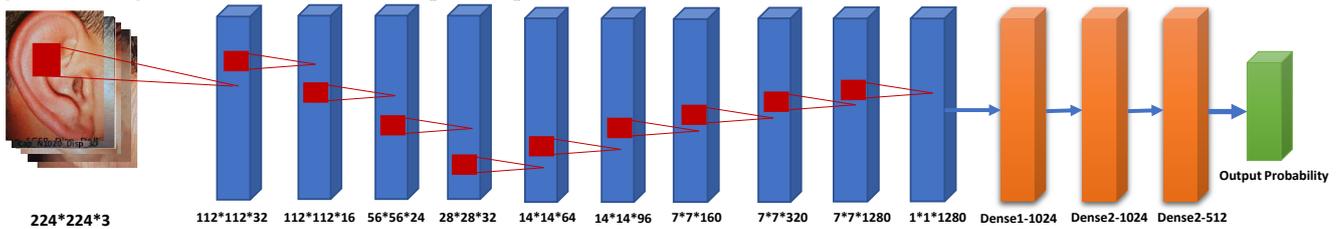

Fig. 5: PADNet architecture. The top layers of MobileNetV2 are changed to execute the binary classification for the presentation attack detection task. The layers highlighted in orange represent the added top layers.

Table 3: Transfer learning parameters on PADNet-1

| Type/Stride | Statues | Activation Function |
|---|---|---|
| 1-26 Layers | Frozen | - |
| 27th Layer | Trainable | RELU |
| Dense 1 | 1024 | RELU |
| Dense 2 | 1024 | RELU |
| Dense 3 | 512 | RELU |
| Dense 4 | 2 | Sigmoid |

As another experiment, we changed the frozen layers and parameters to get better results in the PADNet and called PADNet-2. However, we achieved a worse result. The details of PADNet-2 are shown in Table 4.

Table 4: Transfer learning parameters on PADnet-2

| Type/Stride | Filter Shape | Activation Function |
|---|---|---|
| 1-16 Layers | Frozen | Frozen |
| 17-27 Layers | Trainable | RELU |
| Dense 1 | 1024 | RELU |
| Dense 2 | 1024 | RELU |

| Dense 3 | 512 | RELU |
| Dense 4 | 2 | Sigmoid |

## 4.1. Ear pad analysis

In this section, anti-spoofing methods have been evaluated for ear biometric. Five different attack scenarios have been made. They were supplied to a DNN to train a classifier for distinguishing between fake and real ear biometric. With the proposed algorithm the accuracy of classification has been determined. All fake images have been considered as a class called fake class, however, in the test part, they are evaluated separately.

### 4.1.1. Evaluation metrics

According to the ISO standard, Attack Presentation Classification Error Rate (APCER), Bona-Fide Presentation Classification Error Rate (BPCER) are renamed by False Match Rate (FMR) and False Non-Match Rate (FNMR) respectively [15]. Impostor Attack Presentation Match Rate (IAPMR) is the acceptance of presentation attack incorrectly rate which is inherited from False Match Rate in spoof scenario. In consequence of testing PAD systems, APCER and BPCER are defined as;

$$APCER = \frac{1}{N_{PAIS}} \sum_{i=1}^{N_{PAIS}} (1 - RES_i) \quad (1)[16]$$

Where, $N_{PAIS}$ is a quantity of attack presentations and $RES_i$ shows number of success presentation attack it means:

$$\begin{cases} if \quad RES_i = 1 \quad attack\ presentation \\ if \quad RES_i = 0 \quad bona\ fide\ presentation \end{cases}$$

And for BPCER;

$$BPCER = \frac{\sum_{i=1}^{N_{BF}}(1 - RES_i)}{N_{BF}} \quad (2)[17]$$

$N_{BF}$ = the number of Bona fide presentation

In addition, half total error rate (HTER), which are utilized for evaluating biometrics systems [17]. Since the deep neural network output are probabilities, the assumptions are considered $\tau = 0.5$.

In this paper, all presentation attack detection standards and metrics follow ISO/IEC 30107-3 standards. However, there are still no roles for ear biometric.

### 4.1.2. Ear PAD results

PADNet has been split into two models in PADNet-1 we use Adam optimizer, input size 224*224*3 pixels, batch size 32, and 50 epoch iterations. Furthermore, in PADNet-2 we use a slightly different style of parameterization. We froze 50 layers, input size 224*224*3 pixels, batch size 64, and Stochastic gradient descent (with learning rate 0.0001 and momentum 0.9). The loss values for PADNet-1 and PADNet-2 are 0.0040 and 8.7142e-04 respectively and accuracy in training time for PADNet-1 and PADNet-2 are 0.9986 and 0.9999, respectively. Due to having low loss values and high accuracy during training time, the proposed methods do not achieve a better result during test time. The $BPCER$ and $APCER$ are shown in Table 5 and Table 6, respectively, for both models PADNet-1 and PADNet-2 on the different types of attacks.

Table 5: The BPCER for both models PADNet-1 and PADNet-2 on bona fide presentations

| Detection Algorithm | BPCER (%) |
|---|---|
| PADNet-1 | 99.9 |
| PADNet-2 | 99.9 |

Both deep models (PADNet-1 and PADNet-2) have good accuracy when they are classifying bona fide presentation but in attack presentations, PADNet-1 has achieved better results. According to Table 7, we can see differences between results for different capturing devices and displays. Dell-SGA7 was the most difficult attack presentation to detect. The APCER for Dell-SGA7 has the worst values which are 76.74%. Dell-SGS9 with 82.32% ACPER is another attack presentation which detecting this kind of attack is the most difficult. The other attack presentations have been detected with high accuracy. The results illustrate that using different attack presentations has different accuracy and the possible attack presentations may be taken into account in the system.

Our proposed approach outperformed HTER for the PADNet-1 and PADNet-2, deep learning methods, submitted on the WUT-Ear V1.0 database with various attacks, which are illustrated in the Table 7. Our proposed approach achieved an HTER of 0.08% on display attack (display device and recognition device are SAMSUNG C27JG50QQUX monitor and is Samsung Galaxy A7 respectively) in PADNet-1.

Both deep models (PADNet-1 and PADNet-2) have good accuracy when they are classifying bona fide presentation but in attack presentations, PADNet-1 has achieved better results. According to Table 7, we can see differences between results for different capturing devices and displays. Dell-SGA7 was the most difficult attack presentation to detect. The APCER for Dell-SGA7 has the worst values which are 76.74%. Dell-SGS9 with 82.32% ACPER is another attack presentation which detecting this kind of attack is the most difficult. The other attack presentations have been detected with high accuracy. The results illustrate that using different attack presentations has different accuracy and the possible attack presentations may be taken into account in the system.

Table 6: The APCER for both models PADNet-1 and PADNet-2 on different type of attacks

| Display device | Recognition device | Abbreviation | Detection Methods | APCER (%) |
|---|---|---|---|---|
| Dell Ultra Sharp 32 Ultra HD 4K Monitor | Samsung Galaxy A7 | Dell-GA7 | PADNet-1 | 76.74 |
| | | | PADNet-2 | 75.84 |
| | Samsung Galaxy S9 | Dell-GS9 | PADNet-1 | 82.32 |
| | | | PADNet-2 | 74.89 |
| | Nokia Lumia 1020 | Dell-NL1020 | PADNet-1 | 99.39 |
| | | | PADNet-2 | 99.21 |
| SAMSUNG C27JG50QQUX monitor | Samsung Galaxy A7 | S3D-GA7 | PADNet-1 | 99.83 |
| | | | PADNet-2 | 99.66 |
| | Samsung Galaxy S9 | S3D-GS9 | PADNet-1 | 94.48 |
| | | | PADNet-2 | 91.77 |
| | Nokia Lumia 1020 | S3D-NL1020 | PADNet-1 | 98.2 |
| | | | PADNet-2 | 98.2 |
| | | Print-GA7 | PADNet-1 | 97.57 |

| Brother MFC9340CDW - multifunction printer | Samsung Galaxy A7 | | PADNet-2 | 96.97 |

Our proposed approach outperformed HTER for the PADNet-1 and PADNet-2, deep learning methods, submitted on the WUT-Ear V1.0 database with various attacks, which are illustrated in the Table 7. Our proposed approach achieved an HTER of 0.08% on display attack (display device and recognition device are SAMSUNG C27JG50QQUX monitor and is Samsung Galaxy A7 respectively) in PADNet-1.

Table 7: The HTER for both models PADNet-1 and PADNet-2 on different type of attacks

| Display device | Recognition device | Abbreviation | Detection Methods | HTER (%) |
|---|---|---|---|---|
| Dell Ultra Sharp 32 Ultra HD 4K Monitor | Samsung Galaxy A7 | Dell-GA7 | PADNet-1 | 11.63 |
| | | | PADNet-2 | 12.08 |
| | Samsung Galaxy S9 | Dell-GS9 | PADNet-1 | 8.84 |
| | | | PADNet-2 | 12.555 |
| | Nokia Lumia 1020 | Dell-NL1020 | PADNet-1 | 0.30 |
| | | | PADNet-2 | 0.39 |
| SAMSUNG C27JG50QQUX monitor | Samsung Galaxy A7 | S3D-GA7 | PADNet-1 | 0.08 |
| | | | PADNet-2 | 0.17 |
| | Samsung Galaxy S9 | S3D-GS9 | PADNet-1 | 2.76 |
| | | | PADNet-2 | 4.11 |
| | Nokia Lumia 1020 | S3D-NL1020 | PADNet-1 | 0.9 |
| | | | PADNet-2 | 0.9 |
| Brother MFC9340CDW - multifunction printer | Samsung Galaxy A7 | Print-GA7 | PADNet-1 | 1.21 |
| | | | PADNet-2 | 1.51 |

In addition, we also used attack presentations to evaluate presentation attack detection system without using the attack presentations in the PADNet-1 and PADNet-2 training. Despite the fact that Print-SGA7 was not used in the training, the result of presentation attack detection was 97.57% which is acceptable in comparison with other attack presentations results in this paper.

## 5. Conclusions

In this paper, ear recognition and ear PAD are explored by considering mobile devices application and biometric multimodality. The proposed system uses a smartphone's camera to capture ear shape. Subsequently, to make fake or unreal data for attack purpose, different kinds of material has been used. Different presentation attacks are made for ear PAD. According to the achieved APCER values for different presentation attacks, it is found that using different devices with variable quality achieves different results. Fine-tuned deep neural networks achieved very good results (in the best condition 99.83%) on just some of the attacks. Therefore, we need to improve the proposed method with using more data hence it can be considered as a future research. In addition, in this paper data augmentation is being considered another future research for ear PAD.

## Acknowledgments

This paper was supported by the European Union's Horizon 2020 research and innovation programmed under the Marie Skłodowska-Curie grant agreement No 675087.